\documentclass[10pt,twocolumn,letterpaper]{article}

\usepackage{times}
\usepackage{epsfig}
\usepackage{graphicx}
\usepackage{amsmath}
\usepackage{amssymb}

\usepackage[pagebackref=true,breaklinks=true,letterpaper=true,colorlinks,bookmarks=false]{hyperref}

\begin{document}

\title{ Fidelity-Naturalness Evaluation of Single Image Super Resolution}

\author{Xuan Dong\\
UCLA\\
{\tt\small dongxuan8811@gmail.com}
\and
Yu Zhu\\
Northwestern Polytechnical University\\
{\tt\small zhuyu1986@mail.nwpu.edu.cn}
\and
Weixin Li\\
UCLA\\
{\tt\small lwx@cs.ucla.edu}
\and
Lingxi Xie\\
UCLA\\
{\tt\small 198808xc@gmail.com}
\and
Alex Wong\\
UCLA\\
{\tt\small alexw@cs.ucla.edu}
\and
Alan Yuille\\
UCLA\\
{\tt\small yuille@stat.ucla.edu}
}

\maketitle

\begin{abstract}
We study the problem of evaluating super resolution methods. Traditional evaluation methods usually judge the quality of super resolved images based on a single measure of their difference with the original high resolution images. In this paper, we proposed to use both fidelity (the difference with original images) and naturalness (human visual perception of super resolved images) for evaluation. For fidelity evaluation, a new metric is proposed to solve the bias problem of traditional evaluation. For naturalness evaluation, we let humans label preference of super resolution results using pair-wise comparison, and test the correlation between human labeling results and image quality assessment metrics' outputs. Experimental results show that our fidelity-naturalness method is better than the traditional evaluation method for super resolution methods, which could help future research on single-image super resolution.
\end{abstract}

\section{Introduction}
\label{Sec:intro}
We study the problem of evaluating super resolution (SR) methods and the related metrics. SR is important for different applications and practical systems, such as scalable video coding and high resolution (HR) display devices where the source image/video is of low resolution (LR) and SR is required to fit the image resolution of the client or the display device.

We first define fidelity and naturalness. In this paper, fidelity means the pixel-by-pixel difference between two images. Naturalness means the subjective preference of images from human visual system (HVS). We argue that both fidelity and naturalness should be evaluated for SR methods. For applications like scalable video coding, the SR results are desired to preserve basic information of the original images, so the fidelity evaluation is needed. For applications like HR display, final consumers are usually humans and SR results are desired to get high preference by the human visual system, so the naturalness evaluation is important.

Traditional SR evaluation methods compute the difference (i.e. the fidelity) between the original image $A$ and the SR image $B$, and they judge that $B$ is better if it has higher fidelity score. A main observation of this paper is that traditional evaluation methods have limitations for both fidelity and naturalness.

\begin{figure*}
\begin{center}
\includegraphics[width=0.97\textwidth]{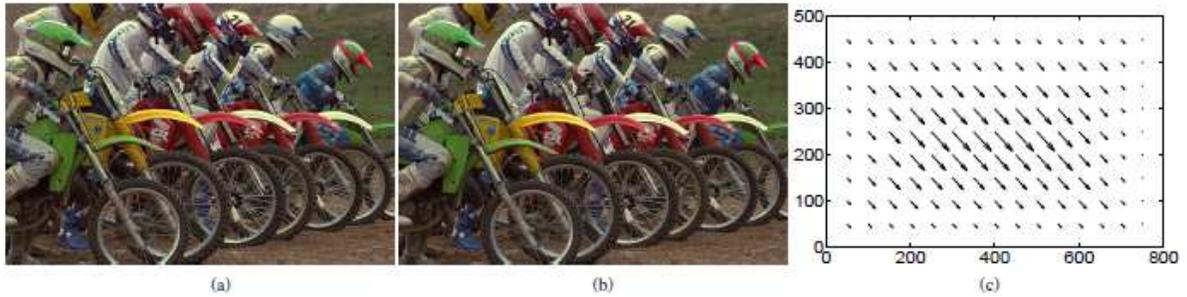}
\end{center}
   \caption{An example of inconsistency between traditional evaluation and naturalness. (a) The original image from the LIVE dataset \cite{LIVE}. (b) The spatially warped image from (a) with the motion vectors visualized in (c). According to traditional evaluation (with PSNR as the metric), the quality of (b) is very low (only 16.79 dB). But both images have almost the same human visual perception. This example shows that the traditional evaluation does not always agree with naturalness.}
   \label{fig:fidelity_example}
\end{figure*}

\begin{figure*}
\includegraphics[width=0.97\textwidth]{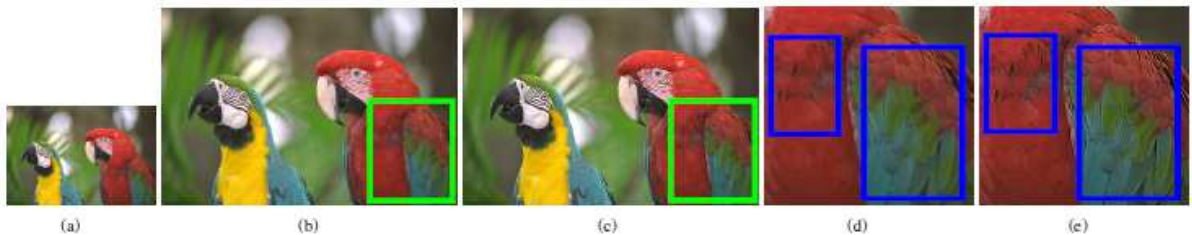}

   \caption{An example showing traditional evaluation is not consistent with either fidelity or naturalness. (a) The low resolution image $a$. (b) The original high resolution image $A$ from the LIVE dataset \cite{LIVE}. (c) The contrast adjustment result $A'$ from $A$. (d) The green box region of (b). (e) The green box region of (c). The blue boxes show the regions with more obvious difference between the two images. According to bilinear down-sampling, both $A$ and $A'$'s down-sampling results are $a$. And due to contrast adjustment, $A'$ has better human visual perception than $A$ at many regions like the blue boxes regions. As we argued in Sec. \ref{sec:fidelity_motivation}, if a super resolution result is similar with $A'$, it should have high fidelity and naturalness values. However, according to traditional evaluation (with PSNR as the metric), $A'$'s quality is very low (only 26.68 dB). This shows traditional evaluation is not always consistent with fidelity and naturalness.}
   \label{fig:quality_example}
\end{figure*}

Firstly, the traditional evaluation result of a SR image is not always consistent with the naturalness of the image, i.e. the preference of HVS. We provide two examples of the inconsistency between traditional evaluation and naturalness in Figs. \ref{fig:fidelity_example} and \ref{fig:quality_example}. In Fig. \ref{fig:fidelity_example}, the original image and the spatially warped image have similar human visual perception, i.e. naturalness. However, according to the traditional evaluation, i.e. fidelity, the quality of the warped image is low. In Fig. \ref{fig:quality_example}, the contrast adjustment image has even higher naturalness than the original image at many regions, but according to traditional evaluation, the quality of the contrast adjustment image is low too.

Secondly, for fidelity evaluation, the traditional evaluation also has limitations. We observed that a set of HR images $\{A\}$ can have the same corresponding LR image $a$ after down-sampling, and the original image $A$ is only one of the images in $\{A\}$. For a SR result $B$, if it is similar with some other image in $\{A\}$, it should also have high fidelity. The reason is that for a SR method, it super resolves the LR image $a$ to $B$. It is impossible for the SR method to judge which image in $\{A\}$ is the original one. In different cases, even for the same LR image $a$, the original image may be different, like the example in Fig. \ref{fig:quality_example}. However, suppose a SR result $B$ is very similar with $A'$ in Fig. \ref{fig:quality_example}, the quality of the image will still be low according to traditional evaluation. This reveals that the traditional method has bias for fidelity evaluation. In short, due to traditional evaluation methods' limitations for fidelity and naturalness, they may mislead the evaluation of SR methods in some cases.

To overcome these limitations, we use both fidelity and naturalness for SR evaluation. For fidelity evaluation, the goal is to test whether a SR image $B$ is similar with some image in $\{A\}$. We proposed a metric to test the difference between image $a$ and image $b$ (the LR images of image $A$ and $B$) as the fidelity, because it greatly reduces the computation cost and can have similar results. For naturalness evaluation, due to the lack of ground-truth preference dataset, we let humans label subjective preference of SR results using pair-wise comparison to get the ground-truth preference.
The human labeling method is good for naturalness evaluation, but, in practical systems, it is difficult to be applied to all images and SR methods due to the large amount of labeling work for humans. The image quality assessment (IQA) metric is a good substitution in practical systems. Thus, we test the correlation between human labeling results and IQA metrics' results, so that we could evaluate and select proper IQA metrics for naturalness evaluation.

The main contributions are summarized as follows.
1. Both fidelity and naturalness are used for SR evaluation.
2. A new metric for fidelity evaluation is designed.
3. For naturalness evaluation, we let humans label the preference of SR results of 8 state-of-the-art SR methods using pair-wise comparison on the LIVE dataset \cite{LIVE}.
4. We test the correlation between IQA metrics (FR and NR) and human labeling results for naturalness evaluation.

\section{Related work}

\subsection{Single image super resolution}
Single image super resolution methods can be summarized into
two categories: the gradient statistical prior based methods
\cite{ZY7,ZY19,ZY20}, and patch example based methods \cite{ZY2,ZY4,ECCV14,SIG11,ZY9,ICCV09,ZY15,TIP10,ZY23,ZY27,CVPR14}.

The gradient statistical prior based methods \cite{ZY7,ZY19,ZY20}
are parametric methods which try to model the gradient
transfer function from low resolution (LR) to high resolution (HR) images. Based on the
fact that sharp edges in the image correspond to concentrated
gradients along the edge, Fattal et al. \cite{ZY7} model the
edge profile distribution. Sun et al. \cite{ZY19} exploit the gradient
transformation from HR image to LR image. And Tai et
al. \cite{ZY20} recover the fine gradient using the user-supplied exemplar
texture.

Patch based methods became popular for the simplicity
to represent the local structure. Patches' self-similar properties are also exploited
in the work \cite{SIG11,ICCV09,ZY14,ZY19}. As in the seminal work of Freeman et al. \cite{ZY9},
Markov Random Field is employed to select the appropriate
HR patch from a bundle of candidates. Chan et al. \cite{ZY2}
proposed a Neighbor Embedding method inspired by LLE
algorithm in manifold learning, followed by extensions of
this work \cite{ZY10,ZY28}. Sparse coding methods \cite{ZY24,TIP10} exploit
the sparsity property in the patch representation. He et al. \cite{ZY13} extend the work by allowing a
mapping function between HR and LR sparse coefficients.
Timeofte et al. \cite{ZY21} also proposed an improved variant of
Anchored Neighborhood Regression (ANR) method. Zhu et al. \cite{CVPR14} allow a patch deformation instead of using patches as a fixed vector, thus making the dictionary more expressive. Cui et al. \cite{ZY4} build a deep network cascade reconstruction structure for super resolution. And Dong et al. \cite{ECCV14} extend the sparse coding method to a
convolutional neural network(CNN) learning structure with
different mapping layers.

\subsection{Image quality assessment (IQA)}

There are two major kinds of IQA methods: full reference (FR) IQA and no reference (NR) IQA metrics.

In general, FR IQA metrics compute the pixel-wise difference between the reference image and the testing image in different transformed fields. The mean squared error (MSE) directly computes two images' mean squared error. Peak signal-to-noise ratio (PSNR) computes two images' MSE in the logarithmic decibel scale. SSIM \cite{SSIM} and UQI \cite{UQI} considers the difference in illumination, contrast, and structure between two images. FSIM \cite{FSIM} considers images' difference in phase congruency and gradient magnitude. IFC \cite{IFC} considers the mutual information of two images in wavelet field. IFC is improved in VIF \cite{VIF}. 


For NR IQA metrics, some learning-based methods \cite{NR3,NR9,NR12,NR13,NR14,NR15,NR16,NR19,NR20,NR21,NR22,NR24,NR25,NR26,NR27} are proposed to simulate HVS to estimate the image quality. In training stage, they learn the relationship between image features and image quality scores using the training set (humans labeled images' quality scores). In the testing stage, they use the learned relationship to estimate the quality score of the test images. They proposed various features for training and testing, like curve-let based feature \cite{NR13}, wavelet based feature \cite{NR14}, DCT based feature \cite{NR15}, and etc. The regression methods they use include support vector regression (SVR) \cite{NR19} \cite{NR12}, neural network based regression \cite{NR20} \cite{NR25}, KL distance \cite{NR21} and etc. This kind of methods can be used in our topic and we perform experiments for the NR IQA metrics like BRISQUE \cite{NR19}, DIVINE \cite{NR12}, BLIINDS2 \cite{NR15}, CORNIA \cite{NR26} and LBIQ \cite{NR14}.

For naturalness evaluation, we test the correlation between IQA metrics' results and labeling results of annotators, because in practical systems the IQA metric is a good substitution for the costly human labeling work.

\subsection{Evaluation of single image super resolution methods}
\label{sec:traditional_evaluation}
Traditional super resolution (SR) evaluation methods usually compute the difference between the original image $A$ and the SR image $B$, i.e. $A$ and $B$'s fidelity, using FR IQA metrics, and use the fidelity results as the quality score of $B$. They assume that higher fidelity will have higher quality. The most complete work, to our knowledge, is Yang et. al.'s benchmark evaluation \cite{ECCV14Benchmark}. To test which metric is proper for SR evaluation, it performs FR IQA metrics and human subject studies, and computes the correlation between FR IQA metrics and human subject studies. The dataset for human subject studies contains 10 images.

Like all the other traditional evaluation methods, Yang et. al. \cite{ECCV14Benchmark} actually evaluate fidelity and use it to judge naturalness. This strategy assumes that 1) the original image $A$ has the highest quality, and 2) the quality of $B$ is determined by its distance with $A$. As explained in Sec. \ref{Sec:intro}, the traditional evaluation method is not always consistent with either naturalness or fidelity. Due to these limitations, in this paper, we proposed a new method by evaluating both fidelity and naturalness for SR images. And a new fidelity evaluation method is proposed due to traditional methods' bias for the original images.


\section{Motivation of fidelity-naturalness evaluation}
\label{sec:framework}
We use both fidelity and naturalness for super resolution (SR) evaluation. In this section, we give more detailed reasons about why the framework is proposed.

\subsection{Motivation of adding naturalness into evaluation besides fidelity}
As mentioned before, fidelity means the pixel by pixel difference between two images.  Naturalness means the subjective preference of images from human visual system (HVS). Although fidelity is widely used in traditional SR evaluation methods, in our opinion, naturalness is also important for SR evaluation because the final consumers are always humans.

In addition, we show that the traditional evaluation method has no direct relationship with the naturalness evaluation by two examples in Fig. \ref{fig:fidelity_example} and \ref{fig:quality_example}. So SR evaluation should also take naturalness into consideration. In Fig. \ref{fig:fidelity_example}, we warp the original image as follows. The motion vector (MV) for pixel $(x,y)$ is set as the minimum value between $\frac{{\min (x,width - x)}}{{width}} \times 40$ and $\frac{{\min (y,height - y)}}{{height}} \times 40$. In other words, at center part, the MV will be large. While at the boundary part, the MV will be small. In Fig. \ref{fig:quality_example}, we enhance the contrast of the original image $A$ to get image $A'$. From the examples, we can see that in Fig. \ref{fig:fidelity_example}, both images have almost the same naturalness from HVS and in Fig. \ref{fig:quality_example}, the contrast adjustment image has even higher naturalness than the original image at many regions. However, according to traditional evaluation (with PSNR as the metric), the warped image and the contrast adjustment image have low quality score because they assume that quality relies on the difference with the original image. This motivates our solution that besides fidelity, naturalness should also be evaluated.

\begin{figure}
\begin{center}
\includegraphics[width=0.48\textwidth]{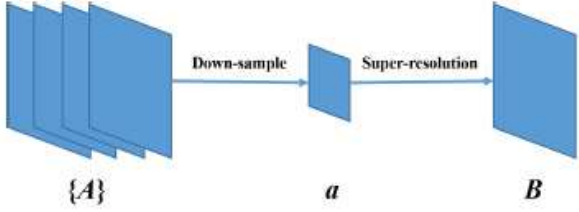}
\end{center}
   \caption{The pipeline of super-resolution. Different from the tradition pipeline, we argue that a low resolution image $a$ has a set of corresponding high resolution images $\{A\}$, instead of only one high resolution image $A$.}
   \label{fig:SR_pipeline}
\end{figure}
\subsection{Motivation of new fidelity evaluation}
\label{sec:fidelity_motivation}
Traditional evaluation methods measure the difference between the original image $A$ and super resolution (SR) result $B$. We argue that, as shown in Fig. \ref{fig:SR_pipeline}, the low resolution (LR) image $a$ has a set of corresponding high resolution (HR) images $\{A\}$. The observation is explained below. When we do down-sampling from the HR image $A$ to the LR image $a$, we can represent it as

\begin{equation}
a = DA,
\end{equation}
where $a$ is an $n \times 1$ vector, $A$ is an $N \times 1$ vector, and $D$ is an $n \times N$ vector which represents the down-sampling process.
Because the rank of $D$ is smaller than $N$, there exist a set of basis $\{e_i\}$ so that $D{e_i}=0$.
Thus, we can get

\begin{equation}
a = D(A + \sum\limits_i {{w_i}{e_i}} ),
\end{equation}
where $w_i$ is const, and the set of $\{A + \sum\limits_i {{w_i}{e_i}}\}$, named $\{A\}$, have the same downsampling result $a$.

Different down-sampling methods have different matrix $D$ in the equation. Thus, the basis $\{e_i\}$ will be also different.
The equation can represent many widely used down-sampling methods like nearest-neighbor, bilinear, bicubic and etc. In case that some down-sampling methods cannot be represented by the equation, however, it does not have an effect on our conclusion that a set of images $\{A\}$ have the same down-sampling result $a$. The reason is that for the set of HR images, the number of images in the whole set is $256^N$. For the set of LR images, the number of images in the whole set is $256^n$. Since each HR image has a corresponding LR image, for each LR image, the number of its corresponding HR images is $256^{N-n}$ on average. The only difference is that for different down-sampling methods, the HR image set $\{A\}$ will be different for the same LR image $a$.

We show an example in Fig. \ref{fig:quality_example} to better explain our observation. In Fig. \ref{fig:quality_example}, $A$ is the original HR image (from LIVE dataset \cite{LIVE}). The LR image $a$ is obtained using bilinear downsampling by a factor of 2. Bilinear downsampling by a factor of 2 computes the mean value for every 4 neighboring pixels and uses it as the value of the corresponding pixel in the LR image. We proposed a simple method to change the contrast of $A$ while not changing the down-sampling result $a$. We first compute the mean value of every 4-neighboring pixels and get the mean-value map $A_m$. Then we can compute the residue $R=A-A_m$, which contains the contrast information. Thus, we can get the contrast enhancement map $E=A+cR$. The result $E$ has the same down-sampling result $a$ with Bilinear down-sampling but has higher contrast than $A$. In Fig. \ref{fig:quality_example}, $A'$ is obtained with $c=4$.

Because a SR method super resolves the LR image $a$ to the SR image $B$. It is impossible for a SR method to judge which image in $\{A\}$ is the original one. In different cases, even for the same LR image $a$, the original image may be different, like the example in Fig. \ref{fig:quality_example}. Thus, for a SR result $B$, if it is similar with any image in $\{A\}$, it should have high fidelity. However, if a SR image $B$ is similar with $A'$ in Fig. \ref{fig:quality_example}, because $dis(A,B)$ is large, the quality of $B$ will be low using the traditional evalution. This explains why traditional methods have bias for fidelity evaluation. So this motivates us to test $dis(\{A\},B)$ instead of $dis(A,B)$ for fidelity evaluation.

\section{Fidelity evaluation}
\label{sec:fidelity}
Our goal of fidelity evaluation is to test the similarity between the super resolution (SR) result $B$ and all images in $\{A\}$. If $B$ is similar with any image in $\{A\}$, it should have high fidelity.

However, because $\{A\}$ has infinite images, it is impossible to test $B$ using each image in $\{A\}$ for fidelity evaluation. We perform down-sampling for $B$ to get $b$ and use the distance between $a$ and $b$ (the low resolution (LR) images of $A$ and $B$) as the fidelity value, i.e. $dis(a,b)$. The reason is that if $B$ is similar with any image in $\{A\}$, $a$ and $b$ will be similar. In addition, if $a$ and $b$ is similar, $B$ must be similar with one image in $\{A\}$. So $dis(a,b)$ can reduce the computation a lot and get similar results.

A direct way for fidelity evaluation is to compute $PSNR(a,b)$. However, we found that different SR methods may assume different down-sampling methods from high resolution (HR) images to LR images. In addition, some SR methods may blur the HR images before down-sampling. Also, some methods may not process the boundary regions of the image due to the lack of neighboring pixels. And the SR images may have misalignment at one or two pixels, in comparison with the original images. In our opinion, for fidelity evaluation, we should prevent the difference of these factors from affecting the final fidelity value, because, for example the choice of down-sampling method, no choice is a wrong one and our evaluation should allow SR methods to use any down-sampling method in their formulation. However, the direct way has to make a choice about the down-sampling method, the blur kernel, the boundary width without computation, and whether the misalignment should be considered. With any choice, the bias may be introduced into the fidelity result.

To avoid the bias, we proposed that for a SR image $B$, the fidelity value is

\begin{equation}
Fd(B) = \mathop {\max }\limits_{i,m,u} PSN{R_c}(a,{f_{i,m,u}}(b)),
\end{equation}
where
\begin{equation}
{f_{i,m,u}}(b) = trans(D{s_m}({k_i}*B),u),
\end{equation}
in the equations, $PSN{R_c}$ is the PSNR value of two images at the center part, with the boundary not computed. The boundary width is set as 20 pixels in this paper. $a$ is the original LR image. $trans$ denotes the translation of the image according to the motion vector $u$. The range of the motion vector is from $[-10,-10]$ to $[10,10]$ in this paper (441 motion vectors in total). $Ds_m$ means the down-sampling process for the HR image according to the $m^{th}$ down-sampling method. In this paper, we include 6 widely used down-sampling methods in Matlab (bicubic, bilinear, nearest-neighbor, box, lanczos2, and lanczos3). $k_i*B$ means the convolution operation for $B$ with the $i^{th}$ kernel $k_i$. In this paper, the kernel size is 3, and the range of the sigma is $[0.1,0.5,0.9,1.3,1.7,2.1]$ (sigma with $0.1$ is equal to no blur for the image). As shown in the equations, from all the PSNR results with different translation, down-sampling and convolution, we select the maximum value as the SR image $B$'s fidelity value.

\section{Naturalness evaluation}
\label{sec:naturalness}
The goal of naturalness evaluation is to obtain the subjective preference of super resolution (SR) images from HVS. Due to the lack of ground-truth (GT) dataset of preference for SR results by human visual system (HVS), we provide a GT dataset.

First, we proposed to label the naturalness of SR results $B$ directly. Second, we use image quality assessment (IQA) metrics to evaluate $B$ and get each metric's result. Last, we compute the correlation between human labeling and IQA metrics' results so that we can test and select the proper IQA metrics for naturalness evaluation.

In image labeling with HVS, we use pair-wise comparison instead of difference mean opinion score (dmos) which is widely used in related work \cite{ECCV14Benchmark} \cite{LIVE} \cite{TID2008} \cite{TID2013}. The reason is that in dmos, people give a score for an image one by one. At different time, people may have different evaluation standards. As a result, two similar images or even the same image may get different dmos scores at different time. This bias can be hardly avoided because people will get tired during labeling and the standard will change gradually. Thus, we propose to use pair-comparison for labeling. At each labeling, the two images are shown together, and people only need to select which has better naturalness. This will decrease the bias a lot. In addition, when the two images are very similar, our software enables to show them one by one at the same position. This makes it easier for people to make the decision.

In labeling, we first get several annotators' results. For each image, we use the labeling results to vote the majority preference of humans. It is because for an image, different people may have different preference. Here, we just use the majority as the GT preference of HVS. Thus, we get the first-step GT preference of HVS.

Detecting outlier annotator is important because several outlier annotators may disturb the GT results. In our work, we compute the correlation between the first-step GT preference and each annotator' preference. If the correlation is below $70\%$, we see it as outlier annotator and remove its labeling data. For the rest of the annotators, we again compute the second-step GT preference of HVS using the same method. This is our final GT preference of HVS.

IQA metrics are desired to simulate HVS for naturalness evaluation in practical applications, because human labeling is costly while IQA metrics can be easily used for evaluating a large amount of results of different SR methods. The estimated naturalness by IQA metrics should be consistent with the naturalness from HVS. To verify the correlation between IQA metrics' results and the subjective preference for pairs of SR results from HVS, we use IQA metrics to produce their preference for each pair. Since IQA metrics usually estimate a score for a given image, to get their preference between two SR images, for each pair, first, we compute the left and right images' quality scores according to the IQA metric. Second, according to the scores, we select the image with higher score as the preferred image of the metric.



\section{Experimental results}
\label{sec:experiments}
\subsection{Experimental settings}
\textbf{Dataset:} We perform our evaluation using the LIVE dataset \cite{LIVE} with 29 images in total. To get low resolution (LR) images, we perform bicubic down-sampling for these images by a factor of 3. Then, we perform super resolution (SR) by a factor of 3 for these LR images with 8 different SR methods. The SR methods include bicubic, Dong et al.'s algorithm \cite{ECCV14}, Freedman et al.'s algorithm \cite{SIG11}, Glasner et al.'s algorithm \cite{ICCV09}, Huang et al.'s algorithm \cite{CVPR15H}, Yang et al.'s algorithm \cite{TIP10}, Zhu et al. 2014's algorithm \cite{CVPR14}, and Zhu et al. 2015's algorithm \cite{CVPR15Z}. For the extern example-based methods, including the methods in \cite{ECCV14}, \cite{CVPR15H}, \cite{TIP10}, \cite{CVPR14} and \cite{CVPR15Z}, the training data is the same as that of Dong et al.'s algorithm \cite{ECCV14}. The SR methods are performed in the Y channel of the YCbCr color space, and all of the results of fidelity and traditional evaluation are obtained by only computing the Y channel's values of the images. The SR results contain 232 images in total. For each image in LIVE dataset, there are 8 SR results and we get 28 pairs of SR results from them. In total, there are 812 pairs of SR results in the dataset.

\textbf{Procedure of labeling:} We invite 23 volunteers to perform pairwise comparison for each pair of SR results. For each pair, the annotators will choose their preference. To help them see the details, our software can show the two images of a pair one by one at the same place. They can view each pair back and forth. The order of pairs and the image order within each pair are random and unknown to annotators so as to avoid bias. The labeling is conducted in the same environment (monitor and room). The outlier annotator detection method is described in Sec. \ref{sec:naturalness}.

\textbf{Image quality assessment (IQA) metrics:} The IQA metrics include (full reference) FR and (no reference) NR IQA metrics. FR metrics include PSNR, SSIM \cite{SSIM}, VIFP \cite{VIF}, UQI \cite{UQI}, and IFC \cite{IFC}. NR metrics include BRISQUE \cite{NR19}, DIVINE \cite{NR12}, BLIINDS2 \cite{NR15}, CORNIA \cite{NR26} and LBIQ \cite{NR14}. To get FR IQA metrics' results, for each SR image, we use the corresponding original image as the reference image.


\subsection{Results}

\begin{table*}[]
\centering
\begin{tabular}{|c|c|c|c|}
\hline
Super resolution method & Fidelity (dB) & Number of preference from HVS & Traditional evaluation (dB) \\ \hline
bicubic   & 48.82         & 1                 & 25.83                       \\ \hline
Dong et al. \cite{ECCV14}    & 45.76         & 113               & 26.22                       \\ \hline
Freedman et al. \cite{SIG11}     & 35.56         & 106               & 24.93                       \\ \hline
Glasner et al. \cite{ICCV09}    & 46.64         & 175               & 26.31                       \\ \hline
Huang et al. \cite{CVPR15H}   & 70.17         & 157               & 26.48                       \\ \hline
Yang et al. \cite{TIP10}     & 48.95         & 74                & 26.17                       \\ \hline
Zhu et al. 2014 \cite{CVPR14}    & 48.62         & 48                & 26.15                       \\ \hline
Zhu et al. 2015 \cite{CVPR15Z}   & 45.79         & 138               & 26.32                       \\ \hline
\end{tabular}
\caption{The fidelity results, number of preference from human visual system (HVS), i.e. naturalness results, and traditional evaluation results (with PSNR as the metric) of the 8 super resolution methods' results on LIVE dataset \cite{LIVE}. The results show that fidelity and naturalness (preference of HVS) is not consistent (like the inconsistency between Freedman et al.'s algorithm and bicubic). And traditional evaluation results are not consistent with either fidelity (like the inconsistency between Glasner et al.'s algorithm and Yang et al.'s algorithm) or naturalness (like the inconsistency between Freedman et al.'s algorithm and Zhu et al. 2014's algorithm).}
\label{tab:fidelity_HVS_traditional}
\end{table*}

The fidelity results, number of preference from HVS (human visual system), i.e. naturalness results, and traditional evaluation results of SR methods' results are shown in Table \ref{tab:fidelity_HVS_traditional}.

As shown, in fidelity results, except for Freedman et al.'s algorithm \cite{SIG11}, all the other SR methods' results get high fidelity values. Huang et al.'s algorithm \cite{CVPR15H} even gets 70.17 dB. Although their fidelity values demonstrate that their SR images' down-sampling versions are not exactly the same as the original LR image $a$, their fidelity is high enough and has little difference with the LR image from HVS.  Freedman et al.'s algorithm is low in fidelity because it enhances the images a lot. As a result, the results are quite different from the original images.

In naturalness results, the methods in \cite{ICCV09}, \cite{CVPR15H}, \cite{CVPR15Z} get the most preference, followed by the methods in \cite{ECCV14} and \cite{SIG11}. Then come the algorithms in \cite{TIP10} and \cite{CVPR14}. Bicubic gets almost no preference. Bicubic's results are not welcome because little repairment for the high-frequency components of the images is done. The results of \cite{ICCV09} , \cite{CVPR15H}, and \cite{CVPR15Z} are very welcome because their edges are more sharp and the contents are more clear than the others.

We can note the results of SR methods are not consistent in fidelity and naturalness, revealing that there is no direct relationship between fidelity and naturalness, and the performance of SR methods in fidelity and naturalness should be tested separately. For example, we can note Freedman et al.'s algorithm. Although its fidelity value is the lowest one among the 8 SR methods, it performs well in human visual perception's labeling and is higher than bicubic, Yang et al.'s algorithm \cite{TIP10} and Zhu et al. 2014's algorithm \cite{CVPR14}, because it can produce more clear edges and content than the other methods for some images, like Fig. \ref{fig:PSNR_naturalness}.

In addition, we can find that traditional evaluation results have no direct relationship with either fidelity or preference of HVS (naturalness). For example, Glasner et al.'s algorithm \cite{ICCV09} is lower than Yang et al.'s algorithm \cite{TIP10} and Zhu et al. 2014's algorithm \cite{CVPR14} in our fidelity results. But in traditional evaluation results, Glasner et al.'s algorithm is higher than Yang et al.'s algorithm and Zhu et al. 2014's algorithm. It is because traditional fidelity computes the difference between the original image $A$ and the SR result $B$, but our fidelity computes the difference between the LR images $a$ and $b$ of $A$ and $B$. Also, Freedman et al.'s algorithm \cite{SIG11} is higher than Zhu et al. 2014's algorithm \cite{CVPR14} in naturalness evaluation. But in traditional evaluation results, Freedman et al.'s algorithm is lower than Zhu et al. 2014's algorithm. The reason of inconsistency between the traditional evaluation method and naturalness is that the difference between $A$ and $B$ has no direct relationship with the naturalness of $B$.

\begin{figure}
\begin{center}
\includegraphics[width=0.23\textwidth, height=0.17\textwidth]{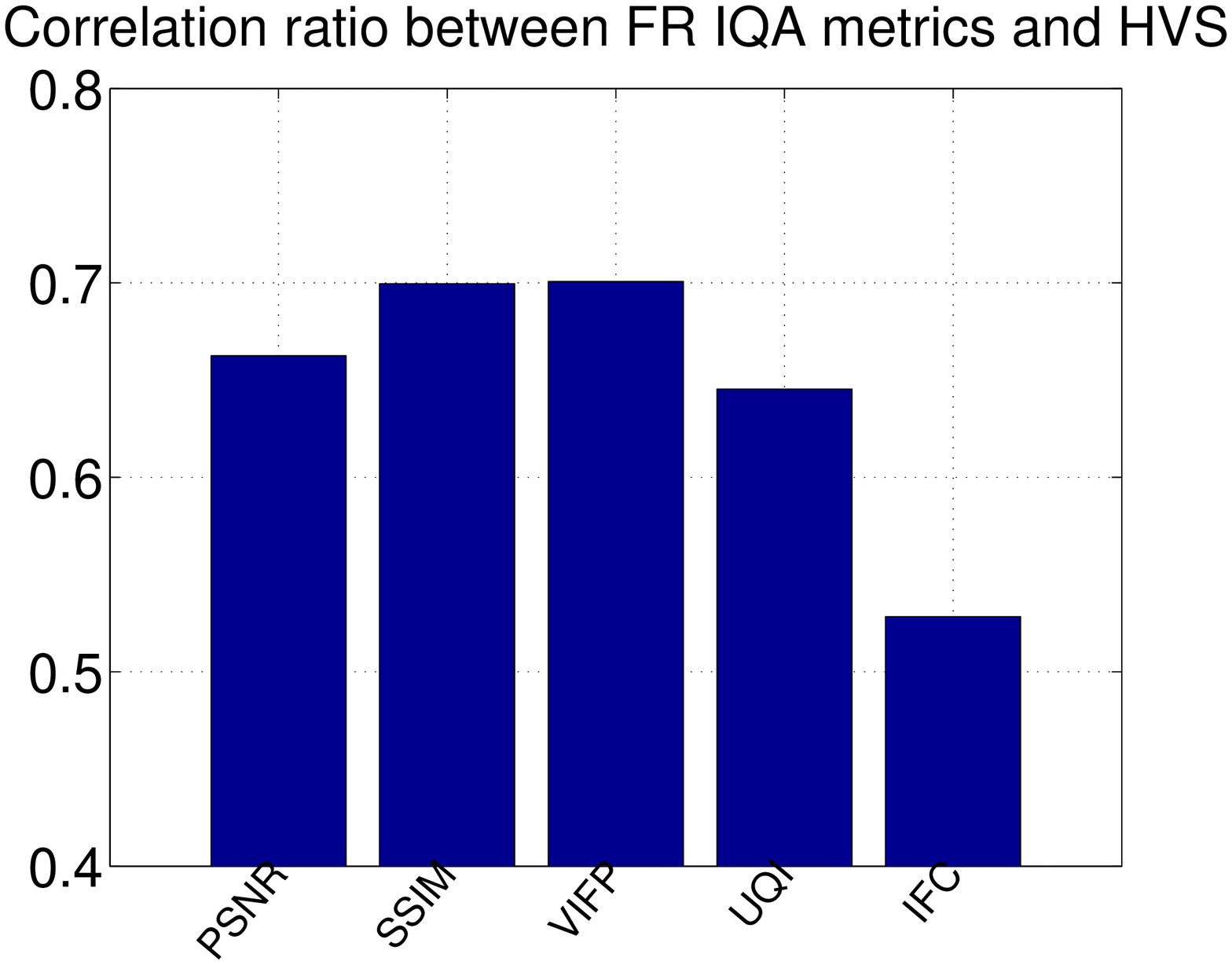}
\includegraphics[width=0.23\textwidth, height=0.17\textwidth]{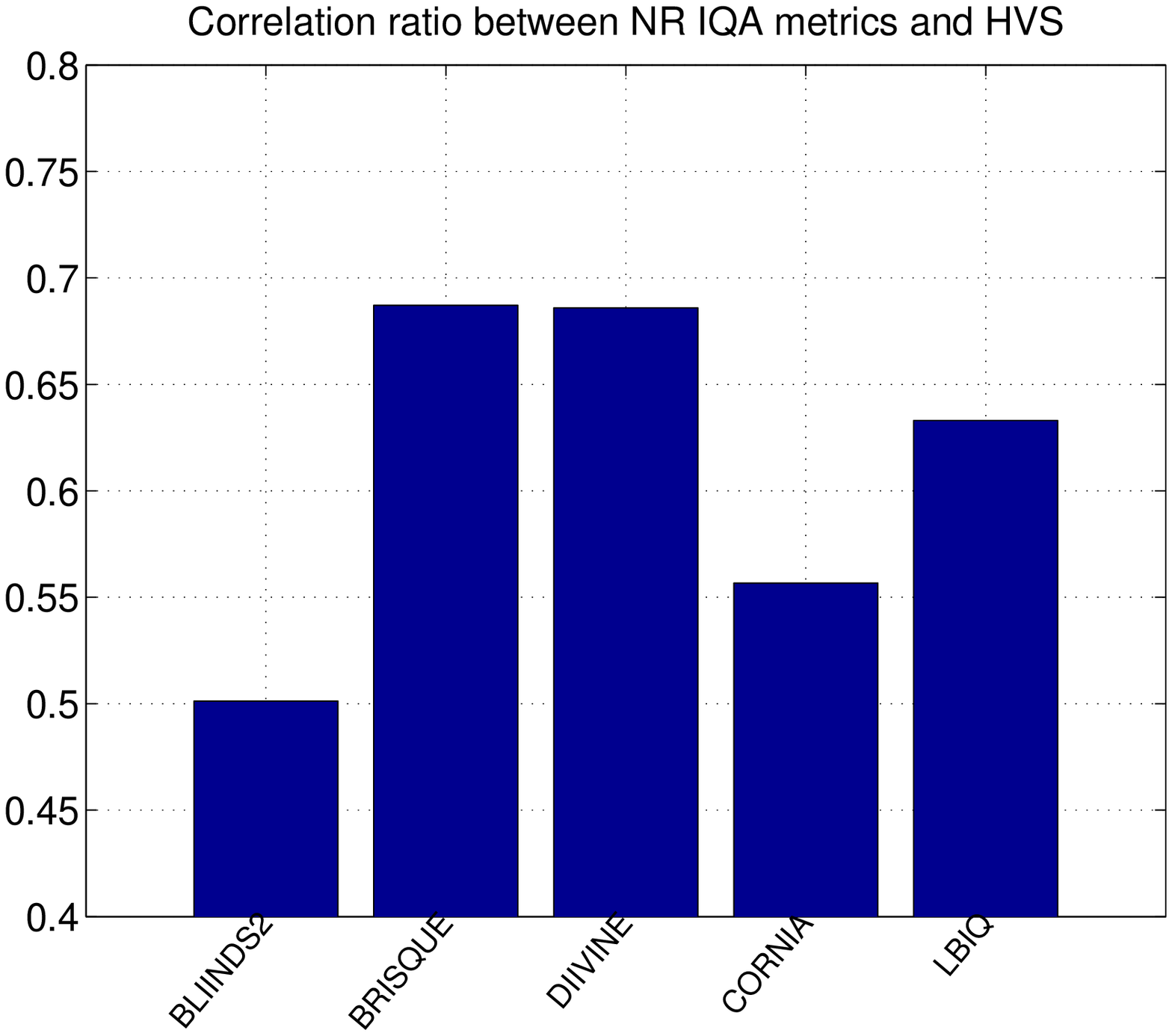}
\end{center}
\vspace{-0.1in}\scriptsize
{\hspace{0.2\linewidth} (a) \hspace{0.5\linewidth} (b)}

%

   \caption{The correlation between preference results of image quality assessment (IQA) metrics and human visual system (HVS). (a) The correlation of full reference (FR) metrics and HVS. (b) The correlation of no reference (NR) metrics and HVS. In FR metrics, SSIM \cite{SSIM} and VIFP \cite{VIF} are the best ones. In NR metrics, BRISQUE \cite{NR19} and DIIVINE \cite{NR12} are the best ones. On the whole dataset, the best FR metric and the best NR metric have similar performances.}
\label{fig:correlation_IQA_HVS}
\end{figure}

For naturalness evaluation, IQA metrics can be a good substitution because human labeling is costly in practical applications.
In Fig. \ref{fig:correlation_IQA_HVS}, we show the correlation between preference results of IQA metrics and HVS. In total, on the whole dataset, FR and NR metrics have similar performance. Because NR metrics do not have to use original images for evaluation, they may be more welcome in practical applications. One reason that FR and NR metrics get similar performance is that most of the SR methods in our experiments have good fidelity. Thus, although FR metrics cannot work well when the fidelity is low, it does not affect FR metrics' results a lot on our dataset.

\begin{figure}
\begin{center}
\includegraphics[width=0.23\textwidth, height=0.17\textwidth]{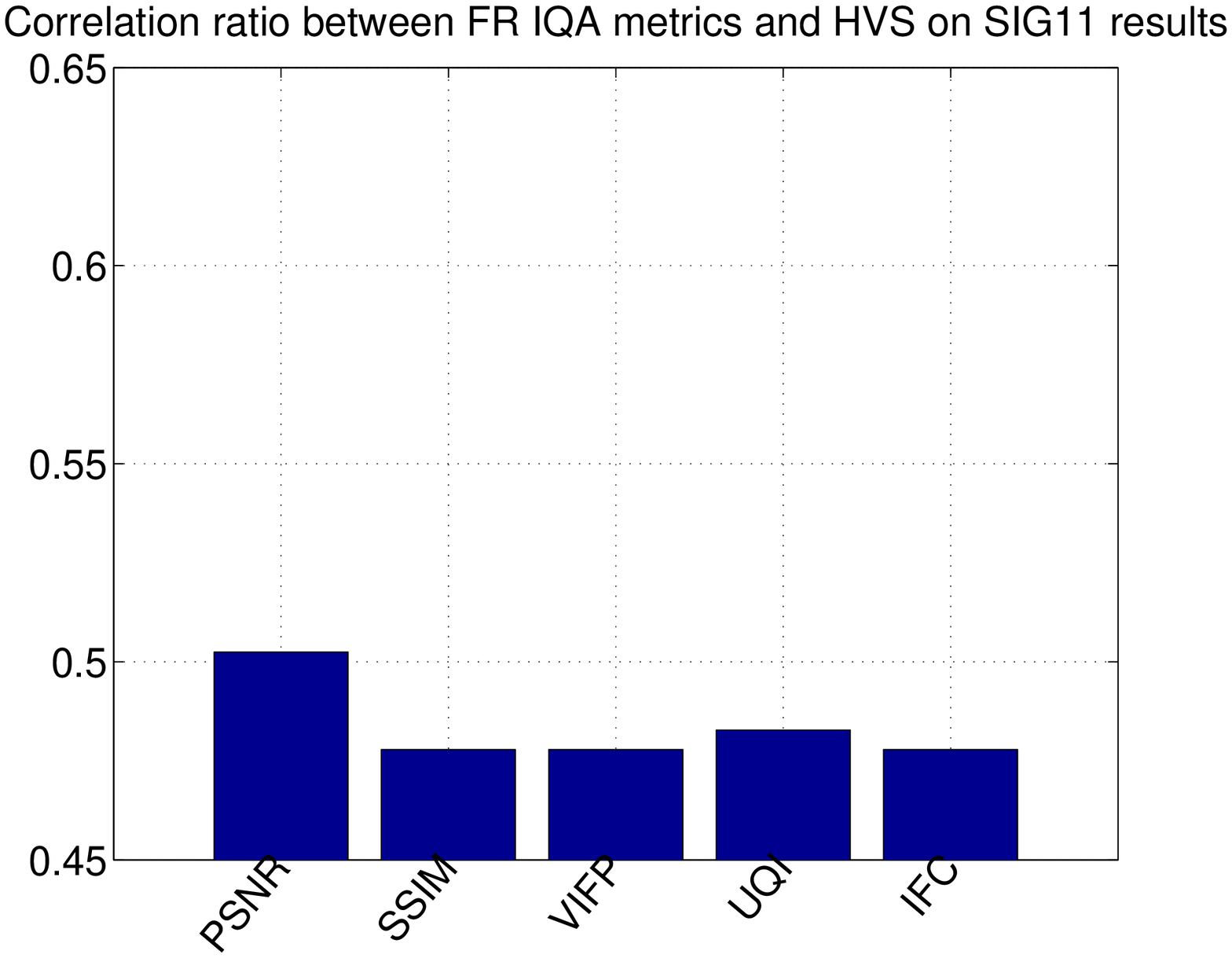}
\includegraphics[width=0.23\textwidth, height=0.17\textwidth]{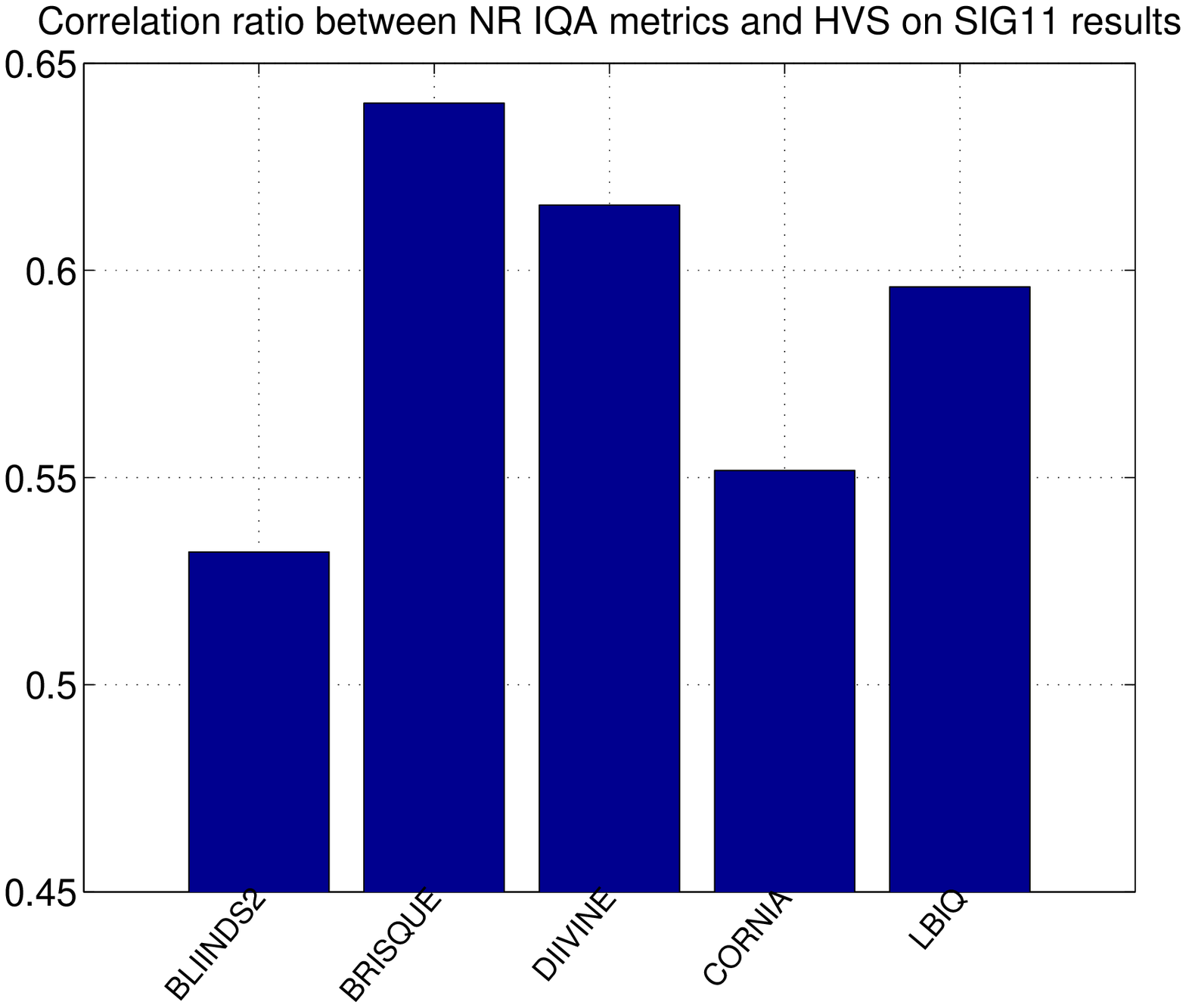}
\end{center}
\vspace{-0.1in}\scriptsize
{\hspace{0.2\linewidth} (a) \hspace{0.5\linewidth} (b)}


   \caption{The correlation between preference of IQA metrics and HVS on Freedman et al.'s results \cite{SIG11}. (a) The correlation of FR metrics and HVS. (b) The correlation of NR metrics and HVS. NR metrics' performances have no big changes comparing with the performances on the whole dataset. But FR metrics' performances have big decrease comparing with the performances on the whole dataset, due to low fidelity of the results of Freedman et al.'s algorithm. This shows NR metrics are more robust than FR metrics for naturalness evaluation.}
\label{fig:correlation_IQA_HVS_sig11}
\end{figure}

To explain it, we select out the pairs which contain the results of Freedman et al.'s algorithm \cite{SIG11}, including 203 pairs in total. We then compute the correlation on this subset. The results are shown in Fig. \ref{fig:correlation_IQA_HVS_sig11}. As shown in the figure, the performance of NR metrics has no big changes, but FR metrics' performance becomes much lower. The reason is that Freedman et al.'s algorithm is low in fidelity, so FR metrics prefer not to selecting the results of Freedman et al.'s algorithm as the better one. However, from HVS, Freedman et al.'s results are welcome for many pairs. This example shows that FR IQA metric is not proper for naturalness evaluation, especially when the practical systems are very complicated that during SR, there may be some other enhancement modules. To sum up, in our opinion, NR IQA metrics are more robust for naturalness evaluation.

\subsection{More examples}

\begin{figure}

\includegraphics[width=0.48\textwidth]{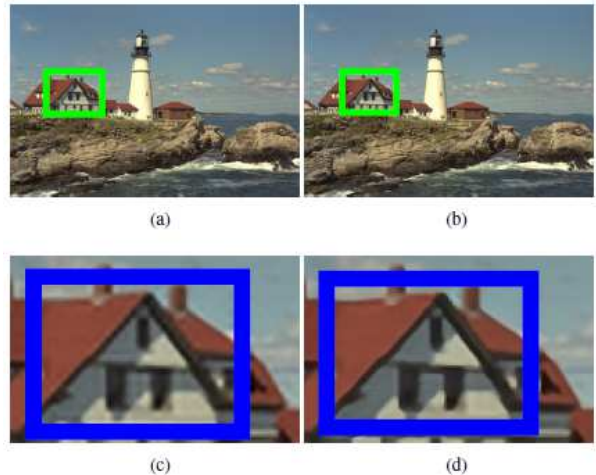}

   \caption{An example showing the inconsistency between fidelity and naturalness. (a) A SR result of Yang et al.'s algorithm \cite{TIP10}. (b) A SR result of Zhu et al. 2015's algorithm \cite{CVPR15Z}. (c) The green box region of (a). (d) The green box region of (b). The blue boxes show the regions with more obvious difference between the two images. The result of Yang et al.'s algorithm (a) has higher fidelity than the result of Zhu et al. 2015's algorithm (b) (51.09 dB vs. 46.94 dB). However, from human visual system, the result of Zhu et al. 2015's algorithm has higher preference.}
\label{fig:fidelity_naturalness}
\end{figure}

In Fig. \ref{fig:fidelity_naturalness}, we show an example of the inconsistency between fidelity and naturalness. Among the total of 812 pairs in our labeling dataset, 447 pairs are not consistent between fidelity and naturalness. As shown, the left image (the result of Yang et al.'s algorithm \cite{TIP10}) has higher fidelity than the right image (the result of Zhu et al. 2015's algorithm \cite{CVPR15Z}) (51.09 dB vs. 46.94 dB). However, from HVS labeling, the right image has higher preference. It is because fidelity and HVS focus on different aspects of the SR images. HVS always focuses on whether the edges are sharp, and the contents are clear. However, fidelity focuses on the pixel-wise difference between the original and SR images.

\begin{figure}
%
\includegraphics[width=0.48\textwidth]{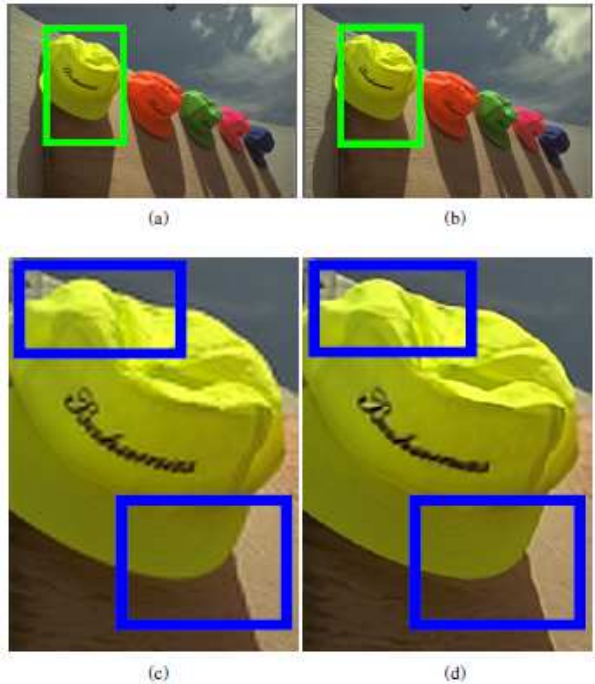}

   \caption{An example showing the inconsistency between traditional evaluation (with PSNR as IQA metric) and naturalness. (a) A SR result of Bicubic. (b) A SR result of Freedman et al.'s algorithm \cite{SIG11}. (c) The green box region of (a). (d) The green box region of (b). The blue boxes show the regions with more obvious difference between the two images. The Bicubic's result has a higher value than the result of Freedman et al.'s algorithm according to the traditional evaluation (31.84 dB vs. 31.17 dB). However, from human visual system, the result of Freedman et al.'s algorithm has higher preference.}
\label{fig:PSNR_naturalness}
\end{figure}

In Fig. \ref{fig:PSNR_naturalness}, we show an example of the inconsistency between traditional evaluation (with PSNR as IQA metric) and naturalness. Among the labeling dataset, 277 pairs are not consistent between traditional evaluation and naturalness. As shown, the left image (Bicubic's result) has a higher value than the right image (the result of Freedman et al.'s algorithm \cite{SIG11}) according to traditional evaluation (31.84 dB vs. 31.17 dB). However, from HVS labeling, the right image has higher preference because it has more clear edges. The reason of the inconsistency is that traditional evaluation computes the difference between the original and SR images. But naturalness is determined by subjective preference of HVS.

\begin{figure}
%
\includegraphics[width=0.48\textwidth]{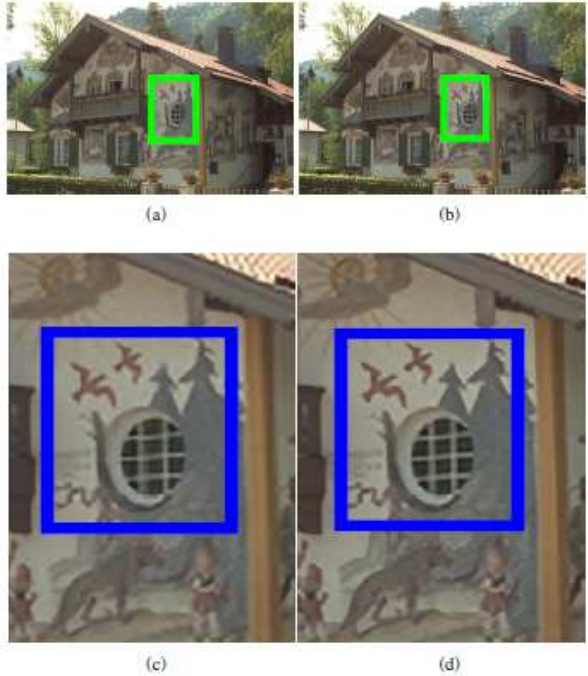}

   \caption{An example showing the inconsistency between traditional evaluation (with PSNR as IQA metric) and fidelity.(a) A SR result of Dong et al.'s algorithm \cite{ECCV14}. (b) A SR result of Glasner et al.'s algorithm \cite{ICCV09}. (c) The green box region of (a). (d) The green box region of (b). The blue boxes show the regions with more obvious difference between the two images. Dong et al.'s result has lower fidelity than Glasner et al.'s result (45.75 dB vs. 46.34 dB). However, from traditional evaluation, Dong et al.'s result has a higher value than Glasner et al.'s result (24.80 dB vs. 24.78 dB).}
\label{fig:fidelity_PSNR}
\end{figure}

In Fig. \ref{fig:fidelity_PSNR}, we show an example of the inconsistency between fidelity and traditional evaluation (with PSNR as IQA metric). Among the labeling dataset, 337 pairs are not consistent between fidelity and traditional evaluation. The reason is that our fidelity computes the difference between the LR images $a$ and $b$, while traditional fidelity computes the difference between the HR images $A$ and $B$ which will have bias for the original image $A$. Because $a$'s corresponding HR image set $\{A\}$ have infinite number of images which may be quite different from each other, the difference between $B$ and $A$ has no direct relationship between the difference between $b$ and $a$.

\section{Conclusions}
In this paper, a new super resolution evaluation method is proposed, which uses both fidelity and naturalness evaluation. In fidelity evaluation, a new metric is proposed due to the bias of traditional evaluation methods. In naturalness, we let humans label preference of super resolution results using pair-wise comparison and obtain a ground-truth dataset of the preference of human visual system. The results show that the traditional evaluation method has bias for both fidelity and naturalness evaluation. In addition, the correlation of human labeling's results and image quality assessment metrics' results are tested, revealing that no reference image quality assessment metrics are more robust than full reference image quality assessment metrics for naturalness evaluation.

{\small
\bibliographystyle{IEEEtran}
\bibliography{egbib}
}

\end{document}